\documentclass[letterpaper, 10 pt, conference]{ieeeconf}  

\IEEEoverridecommandlockouts                              

\usepackage{url,graphicx,tikz,amsmath,bbm,booktabs,array,algorithm,algpseudocode,indentfirst}
\usepackage{amsfonts,dsfont,bm,gensymb}
\usepackage[colorlinks,linkcolor=red]{hyperref}

\title{\LARGE \bf
Attention-Oriented Action Recognition for Real-Time Human-Robot Interaction
}

\author{Ziyang Song$^{1}$, Ziyi Yin$^{1}$, Zejian Yuan$^{1}$, Chong Zhang$^{2}$, Wanchao Chi$^{2}$, Yonggen Ling$^{2}$, and Shenghao Zhang$^{2}$
\thanks{$^{1}$Institute of Artificial Intelligence and Robotics, Xi'an Jiaotong University, Xi'an, China}%
\thanks{$^{2}$Tencent Robotics X, Shenzhen, China}%
\thanks{*Correspondance: {\tt\small songzy305@yahoo.com}}
}

\begin{document}

\maketitle
\thispagestyle{empty}
\pagestyle{empty}

\begin{figure*}[t]
    \centering
    \includegraphics[width=1.0\linewidth]{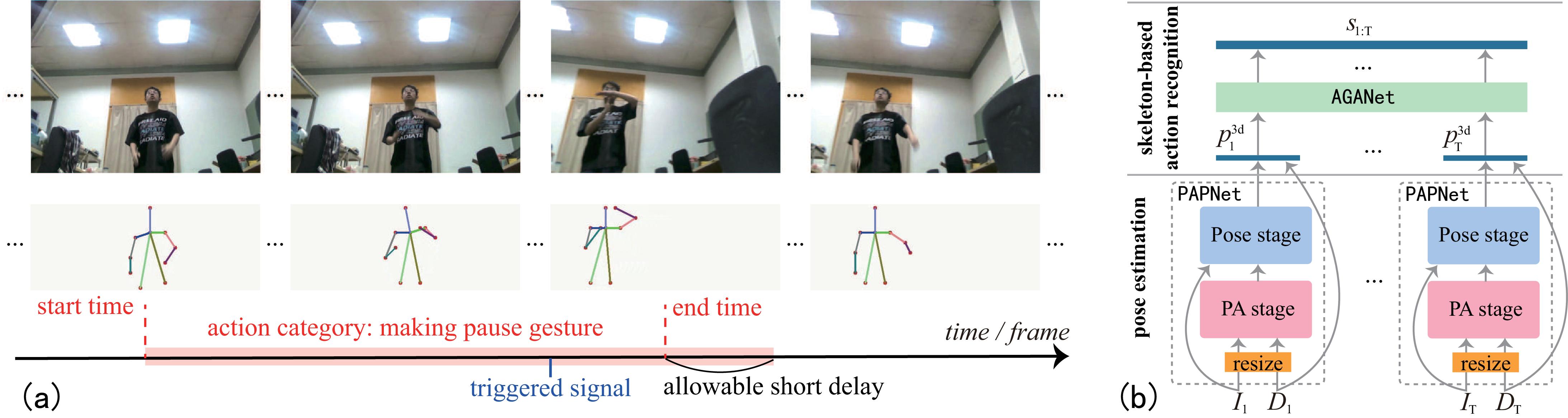}
    \caption{
    \textbf{(a) The action recognition task on a continuous stream in HRI and (b) the proposed framework to solve the task.}
    Within the duration of each action (short delay is allowed), the algorithm is required to trigger a signal, informing the appearance of an action instance of the specific category.
    The person in the scene presents to interact with the robot behind the camera sensor.
    The camera viewpoint is changing because of the robot's actions to respond to the interactor.
    }
    \label{figurelabel}
\end{figure*}

\begin{abstract}

Despite the notable progress made in action recognition tasks, not much work has been done in action recognition specifically for human-robot interaction.
In this paper, we deeply explore the characteristics of the action recognition task in interaction scenarios and propose an attention-oriented multi-level network framework to meet the need for real-time interaction.
Specifically, a Pre-Attention network is employed to roughly focus on the interactor in the scene at low resolution firstly and then perform fine-grained pose estimation at high resolution.
The other compact CNN receives the extracted skeleton sequence as input for action recognition, utilizing attention-like mechanisms to capture local spatial-temporal patterns and global semantic information effectively.
To evaluate our approach, we construct a new action dataset specially for the recognition task in interaction scenarios.
Experimental results on our dataset and high efficiency (112 fps at 640 $\times$ 480 RGBD) on the mobile computing platform (\emph{Nvidia Jetson AGX Xavier}) demonstrate excellent applicability of our method on action recognition in real-time human-robot interaction.

\end{abstract}

\section{Introduction and Related Work}

Human action recognition has long been one of the most popular research topics in computer vision and intelligent robotics.
Its research results are widely used in various applications such as surveillance, healthcare monitoring and human-robot interaction~\cite{survey_application}.
In recent years, large scale video datasets like Sports-1M~\cite{sports1m}, Kinetics~\cite{kinetics}, ActivityNet~\cite{activitynet} and THUMOS14~\cite{thumos14} are proposed, covering rich scenarios and action categories.
PKU-MMD~\cite{pkummd} and NTU RGB+D~\cite{ntu_rgbd} further provide multi-modality data (RGB, depth and skeleton joint coordinates).
With these datasets and the introduction of deep learning, significant progress has been made in action recognition~\cite{survey_deepaction}.

However, as one of its core applications, human-robot interaction (HRI) cannot directly benefit from such progress.
On the one hand, HRI systems are usually embedded in mobile robot platforms which are often limited in computational resources.
Therefore, the state-of-the-art action recognition methods~\cite{soa_class1, soa_detect1, soa_detect2} trained on those large-scale datasets are too computationally intensive to adopt.
On the other hand, data collected in interaction scenarios differ from those datasets for general action recognition tasks.
Thus datasets and metrics specifically for interaction scenarios are needed.

Unlike general action recognition tasks that aim to either classify a segmented clip or classify and meanwhile temporally localize actions from an unsegmented sequence in an offline manner, this task intends to trigger a signal online for each action encountered in a continuous stream.
The output form of triggered signals is similar to online action detection~\cite{online1, online2, online3} or early event detection~\cite{early1}, while our task further specifies scenario and computing platform limitations.
Fig. 1(a) shows an example.
Triggered signals can be directly used in HRI system to guide robots to make instant responses.

In interaction scenarios, people's actions are mostly related to their body movements rather than surrounding environments.
Therefore, skeleton-based action recognition methods should be adopted, given their robustness to illumination change and scene variation~\cite{survey_sktbased}.
In this way, human pose estimation is needed to extract skeleton sequences of interactors from raw videos.
In interaction scenarios, irrelevant people often appear with the interactor.
Single-person pose estimators~\cite{hourglass, cpm} fail to deal with such scenes.
Most of multi-person pose estimation methods fall into two groups: bottom-up and top-down methods.
The former~\cite{singledepth, deepcut, paf} performs estimation for all people in parallel, and the interactor can be determined with extra selection modules.
Yet for HRI, accurate multi-person pose estimation (especially for irrelevant people) at high resolution is excessive consumption of limited computational resources.
The latter~\cite{topdown1, rmpe, topdown2, topdown3} employs a human detector first, and performs single-person pose  estimation for the interactor.
However, selection modules based on human detection results are tough to design, since bounding boxes lack compactness to describe human bodies.
Besides, human detectors also bring considerable waste to some extent on encoding irrelevant people.
Compared to the methods above, applying some rough pre-attention and quickly focusing on the interactor is more feasible.

As for skeleton-based action recognition, spatial-temporal patterns of human actions can be modelled by either RNNs~\cite{sktrnn2, sktrnn3, beyondjoints, skt_va} or CNNs~\cite{skt_va, sktcnn1, sktcnn2, sktcnn4}.
Recently, graph neural network (GNNs)~\cite{stgcn, sktgnn1, sktgnn2, sktgnn3} emerges as a more natural choice to implicitly form a hierarchical representation of the skeleton sequence.
Since the action space in interaction is somewhat simpler than that in general action recognition tasks, directly importing state-of-the-art methods is not reasonable because of the imperfect match between those methods and data in interaction.
A targeted-designed network is needed instead to ensure adaptivity and effectiveness in HRI.

Considering the challenges discussed above, we propose an attention-oriented multi-level network framework to solve this task, as Fig. 1(b) shows.
In the first level, we devise a Pre-Attention Pose Network (PAPNet) for pose estimation in an end-to-end manner.
With Pre-Attention, PAPNet roughly focuses on the interactor in complex changing scenes.
Then computation resources are concentrated on estimating the interactor's pose at high resolution accurately.
In the second level, a compact Attention-Guided Action Network (AGANet) is designed for skeleton-based action recognition.
Two kinds of attention-like mechanisms are incorporated into this model to focus on most important local structures while encoding pose representations from skeleton sequence, and to combine multi-scale temporal motion features to re-allocate representational power of the network before recognition.

Currently, datasets in which subjects appear to interact with a robot behind the camera sensor are still vacant.
In order to verify the effectiveness of our method and facilitate further research on action recognition in HRI, we construct a new multi-modality human action dataset.
We name it as AID (Action-in-Interaction Dataset) since we wish better interaction makes robots a better aid for people's lives.
Within the scope of our knowledge, the AID dataset is the first action recognition dataset to collect from the simulated viewpoints of the mobile robot in HRI.
We also define a new evaluation metric on our dataset.

We deploy the proposed framework on a mobile robot platform embedded with \emph{Nvidia Jetson AGX Xavier} for computing.
Real-time HRI is achieved and demonstrated in the supplementary video.
\emph{Our code and dataset will be made publicly available later.}

The major contributions of our work are summarized as follows:

1) We specify a new action recognition task for HRI, which requires instant responses for actions performed by the interactor.

2) We propose an attention-oriented multi-level network framework, in which multi-granularity attention is integrated for different levels, towards real-time action recognition in interaction scenarios.

3) We construct a new dataset and define a new evaluation metric on it to support further study on the action recognition task in HRI.
Our proposed method achieves superior performance on this new dataset, with also high efficiency to meet real-time requirements for interaction.

\section{Proposed method}

In the following we illustrate the two levels in the proposed framework separately.

\subsection{Pre-Attention Pose Network (PAPNet)}

\begin{figure}[t]
    \begin{center}
        \includegraphics[width=1.0\linewidth]{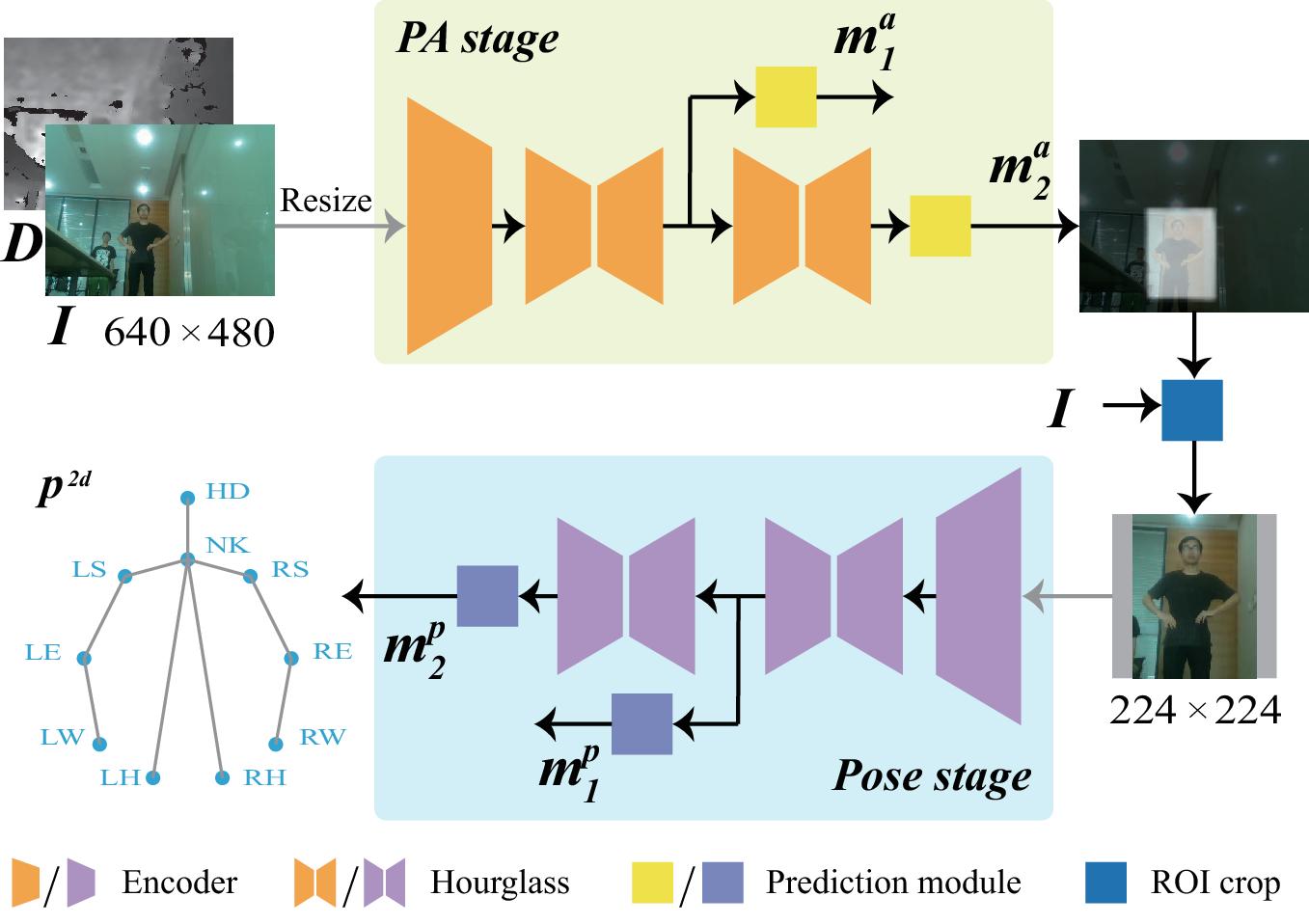}
    \end{center}
    \caption{
    \textbf{Structure of the proposed Pre-Attention Pose Network (PAPNet).}
    10 skeleton joints of the upper body are estimated: head (HD), neck (NK), left/right shoulder (LS/RS), left/right elbow (LE/RE), left/right wrist (LW/RW), and left/right hip (LH/RH).
    }
\end{figure}

Here we propose a compact pre-attention network named Pre-Attention Pose Network (PAPNet), as illustrated in Fig. 2.
PAPNet estimates the 2D pose $p^{2d}$ of the interactor in a multi-person scene from RGB color image $I$ and depth image $D$.
Then given inner-parameters of the camera and depth information, we can project 2D pose $p^{2d}$ back to 3D skeleton $p^{3d}$ in the spatial coordinate system.
According to the actual application requirements, our task focuses on the 10 skeleton joints of the upper body, as Fig. 2 shows.

\begin{figure*}[t]
    \begin{center}
        \includegraphics[width=0.75\linewidth]{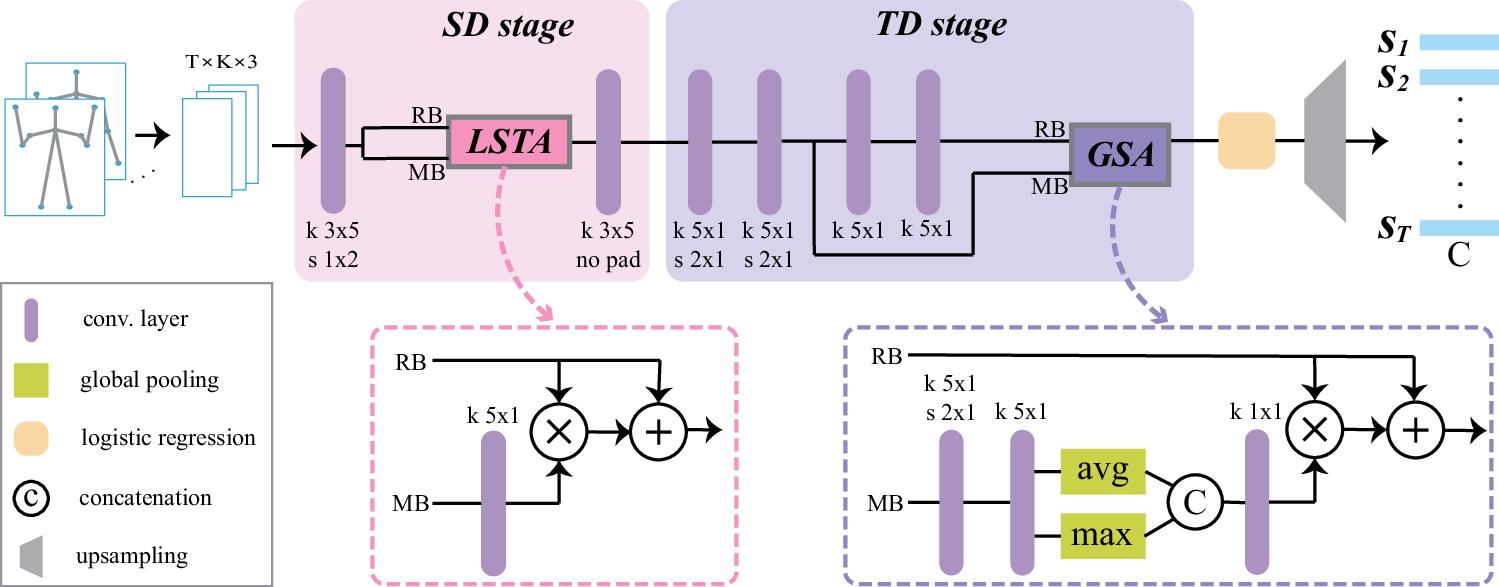}
    \end{center}
    \caption{
    \textbf{An overview of the proposed Attention-Guided Action Network (AGANet).}
    Kernel size and stride of each conv. layer are denoted under itself with "k" and "s".
    The "RB" and "MB" indicate the residual and soft mask branch in each attention module respectively.
    }
\end{figure*}

The PAPNet can be seen as a two-stage model.
In the first Pre-Attention (PA) stage, the low-resolution ($224 \times 224$) RGB color image $I$ and depth map $D$ are integrated as input.
The depth map $D$ provides additional information for localizing the interactor from a complex scene.
PA stage outputs pixel-wise dense attention map $m^{a}$ denoting the rough position of the upper body of the interactor.
Then a local ROI (Region of Interest) suggested by the PA stage is cropped from the original image ($640 \times 480$).
On the ROI, the second Pose stage performs fine-grained single-person pose estimation at high resolution.
Both of the two stages are constructed by stacking two hourglasses~\cite{hourglass} with proper compression.
The output from PA stage is binarized and a minimum bounding rectangle of the largest binary connected component is extracted for ROI crop.

The pixel-wise attention from the PA stage is relatively rough.
Therefore, we can employ shallow network layers in the PA stage and concentrate on local regions that are informative for the overall action recognition task earlier.
As a comparison, top-down or bottom-up approaches spend many resources on encoding more refined bounding boxes or joints of unrelated people.
Furthermore, the PA stage learns to exploit depth information and extract the interactor via the network itself, eliminating hand-designed constraints.
In contrast, both top-down and bottom-up methods need extra selection modules for determining the interactor.

\subsection{Attention-Guided Action Network (AGANet)}

Given a sequence of skeleton joints $p^{3d}_{1:T}$ in the form of 3D coordinates , we arrange it into a $T \times K \times 3$ skeleton image.
Although the sequence length $T$ is usually much longer than the number of joints $K$, we do not resize the skeleton image to a typical image size like $224 \times 224$ as most skeleton-based action recognition methods using CNNs do~\cite{sktcnn1, sktcnn2}.
That's because for interaction scenarios, actions to be encoded are usually very short in time. 
Resizing in that way severely compresses the $T$ dimension and loses discriminative information for actions.
Given the unbalanced aspect ratio of the skeleton image, encoding spatial and temporal patterns simultaneously with typical CNN architectures is not feasible due to a massive gap between the receptive fields needed for the two dimensions.

To overcome such limitations, we propose a novel network named Attention-Guided Action Network (AGANet), with a fully convolutional network (FCN) structure to make dense frame-wise estimation on skeleton sequences.
As shown in Fig. 3, the proposed network is split into two stages:
In the first Space-Dominant (SD) stage, two $3 \times 5$ conv. layers encode relations among skeleton joints in a short term into local spatial-temporal feature representations.
In the second Time-Dominant (TD) stage, there are four $5 \times 1$ conv. layers, of which the first two perform $2 \times$ downsampling operations meanwhile to ensure a longer time interval for subsequent layers to observe the sequence.
Long-term motion patterns are captured in this stage owing to sufficient receptive fields on the $T$ dimension.
Dense estimation on the $T$ dimension is performed at the end to regress scores $s_t$ for actions in each frame.
Two attention-like mechanisms conforming to the idea of residual attention~\cite{resatt} are incorporated, which will be illustrated next.

\textbf{Local Spatial-Temporal Attention (LSTA) module. }
In the sequence, most actions can be distinguished according to the movements of certain joints in specific frames without referring to other spatial-temporal areas.
To make more efficient use of computation resources and representation power of our compact network, we propose a Local Spatial-Temporal Attention (LSTA) module.
After the first conv. layer in the SD stage, spatial-temporal patterns in small local regions have been extracted.
As Fig. 3 shows, in the soft mask branch of the LSTA module, one $5 \times 1$ conv. layer describes the evolutions of each local region in dense short time intervals, hence search for the most important local structures.
Then the calculated attention information guides the next layer in the SD stage to focus on those key local structures while encoding larger regions.

\textbf{Global Semantic Attention (GSA) module. }
For dense estimation of actions in the $T$ dimension, high-level features from deeper layers provide more context and more comprehensive semantic category information, while low-level features better retain frame-wise information.
To combine the advantages of them, we perform frame-wise estimation on low-level features with guidance from high-level features.
Therefore, a Global Semantic Attention (GSA) module is introduced, as illustrated in Fig. 3.
To keep the network compact, we do not increase the depth of the soft mask branch in the GSA module, but perform further downsampling to capture global context information from longer time intervals instead.
After two conv. layers we squeeze the $T$ dimension by a combination of maximum and average pooling to attain rich semantic information of the whole sequence.
Finally, with a $1 \times 1$ conv. layer, high-level features are adjusted channel-wisely to the need of guiding low-level features for feature selection before final estimation.

\subsection{Training procedure}

\textbf{Loss function for PAPNet. }
Following the idea of intermediate supervision~\cite{cpm, hourglass}, the model is trained to repeatedly produce the confidence maps for the locations of Pre-Attention in the PA stage and joints in the Pose stage.
The costs on the output after each hourglass module are added together, resulting in the final loss, i.e.,
\begin{equation}
    L_{A} = {\sum_{t=1}^{2} ||\widehat{m}_{t}^{a} - m_{t}^{a}||_{2}^{2}},
\end{equation}
\begin{equation}
    L_{P} = {\sum_{t=1}^{2} \sum_{k=1}^{K} ||\widehat{m}_{t, k}^{p} - m_{t, k}^{p}||_{2}^{2}},
\end{equation}
where $\widehat{m}$ denotes the groundtruth, $t$ and $k$ index hourglass modules and joints respectively.

\textbf{Loss function for AGANet. }
To allow for batch learning, we evenly sample fixed-length subsequences from each complete sequence to form our training set.
Dense frame-wise category labels are generated according to original annotations, in which category and start and end time of each action instance are annotated.
Frame-wise cross entropy ($CE$) loss is minimized for binary classification on each category, i.e.,
\begin{equation}
    L_{Action} = \frac{1}{T} {\sum_{t=1}^{T} \sum_{c=1}^{C} CE(\widehat{s}_{t, c}, s_{t, c})},
\end{equation}
where $\widehat{s}$ denotes the groundtruth, $t$ and $c$ index frames and action categories respectively.

\textbf{Data augmentation in training AGANet. }
The action recognition in interaction scenarios must maintain the invariance to the camera viewpoint. 
However, even a small part of the space of camera viewpoint changes can never be covered during data collection.
To enrich diversity in the training set and avoid overfitting, we jointly use three data augmentation strategies.
For each fixed-length 3D skeleton subsequence in the training set, 
a) \textbf{rot:} randomly rotate the whole subsequence within $5 \degree$ in the 3D camera coordinate system;
b) \textbf{dist:} randomly adjust the distance from the skeletons to the camera origin by no more than $5\%$;
c) \textbf{gt:} randomly adjust the annotated start and end time of action instances by no more than $5\%$ of the time length of action instances.

Among the three strategies above, \textbf{rot} and \textbf{dist} aim to improve the robustness of recognition to the movements of the camera sensor. 
Meanwhile, \textbf{gt} is committed to reducing the bias among perceptions of different people during annotating.
In this way, transition boundaries between different actions are statistically smoothed and the network can learn to focus on the process of actions.
In every epoch, new data augmentation parameters are randomly generated for each sample.

\subsection{Prediction and post-processing}

During prediction, we firstly extract the interactor's 3D skeleton frame-wisely from a continuous RGBD video stream.
Then we slide fixed-length temporal windows on the skeleton data stream to generate skeleton images and input them into our AGANet to obtain frame-wise category scores.
Since overlaps among different temporal windows exist, we choose the scores from the middle part of each window to form final prediction results on the stream.
Frames without category scores larger than the set threshold are considered as containing no defined actions.
For other frames, the action category with the largest score is granted in each frame.

\begin{figure}[t]
    \begin{center}
        \includegraphics[width=1.0\linewidth]{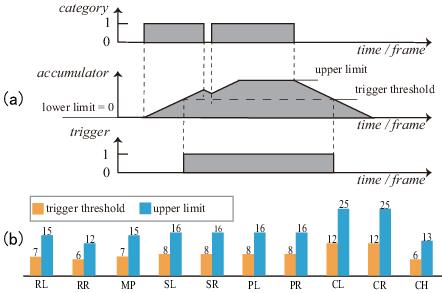}
    \end{center}
    \caption{
    \textbf{(a) Visualization and (b) parameter configurations of post-processing.}
    }
\end{figure}

During post-processing, an accumulator and a trigger are independently set for each action category.
In the sequence, the continuous appearance of a certain action category increases its accumulator score.
When the accumulator score exceeds the trigger threshold, a signal denoting an action instance is triggered.
The trigger state is also changed from $0$ to $1$ to avoid being triggered repeatedly by the same action instance.
When that category no longer appears continuously, its accumulator score gradually decreases and drops below the trigger threshold.
The trigger state is reset to $0$ again and waits to be triggered by the next action instance of this category.
In the accumulator, lower and upper limits ensure the sensitivity, i.e., the accumulator score can rapidly exceed the trigger threshold when the action appears continuously and drop below it otherwise.
Fig. 4(a) visualizes an example of the accumulator score and trigger state of a certain action category influenced by frame-wise estimation over a while in the sequence.
The trigger threshold and upper limit for each category are independently set based on minimum durations of that category of actions, as shown in Fig. 4(b).

\section{Experiments}

\subsection{Dataset and evaluation metrics}

\textbf{Action-in-Interaction Dataset (AID). }
This is our newly collected dataset for the action recognition task in HRI.
Our dataset is captured via the \emph{Intel RealSense D435} camera, which can record RGB color and depth images synchronously.
In interaction scenarios, a robot platform with interaction system is hardly stationary.
So we continuously move the camera sensor while collecting data to simulate the actual situation, leading to video sequences with changing camera viewpoints in the dataset.
We define 10 action categories that are common and conveniently performable in HRI scenarios:
raising left/right hand (RL/RR), making pause gesture (MP), swinging left/right hand (SL/SR), pushing forward with left/right hand (PL/PR), circling with left/right hand (CL/CR), and crossing hands (CH).
We invite 20 subjects and collect $5 \sim 6$ video sequences for each of them.
Each sequence lasts about $60 \sim 80$ seconds (recording with 30 fps) and mostly contains 10 action instances (each defined action category appears once).
Both RGB color and depth images are recorded with $640 \times 480$ resolution.
The total scale of the dataset is 205,138 frames from 102 videos, with 1031 annotated action instances.

\textbf{Cross-subject evaluation. }
We follow the commonly-used cross-subject evaluation~\cite{ntu_rgbd} to split our subjects into training and testing groups, composed of 14 and 6 subjects respectively.
There are 71 videos in the training set and 31 videos in the test set. Such a split setup aims to test the robustness to intra-category variations among different interactors, like body shape and behavioral habit, etc.

\textbf{Metrics. }
We adopt the calibrated average precision ($cAP$)~\cite{online1} to evaluate frame-wise estimation before post-processing in Sec. II.D.
However, triggered signals are the directly expected output form by the task.
Moreover, our post-processing achieves the same function of suppressing false positive frame-wise predictions as $cAP$.
Therefore, we propose a trigger-based metric to evaluate triggered signals.

For each video, the category and trigger time of triggered action instances are recorded.
Based on the idea that an action instance should be discovered between its start and proper delay after its end, we delay the end time of action instances by $20\%$ of their durations in the groundtruth annotations during evaluation.
Then we match a triggered action instance to an annotated one with the same category and count it as a true positive ($TP$) prediction if the trigger time of the former is within the extended duration of the latter.
Triggered actions and annotated actions not successfully matched are denoted as false positive ($FP$) and false negative ($FN$) predictions respectively.
The score threshold for category assignment during prediction can be varied to evaluate the trigger-based average precision ($AP_{trig}$).
We also set the score threshold to $0.4$ to calculate trigger-based precision ($P_{trig}$) and recall ($R_{trig}$).

\subsection{Implementation details}

PAPNet and AGANet are independently trained on an NVIDIA  Tesla M40 GPU.
Although PAPNet can be optimized end-to-end, we find it more efficient to train two stages separately.
Training samples for the Pose stage are cropped from original images according to annotated bounding boxes, with random scaling and rotation for augmentation.
Such a strategy prevents the Pose stage from being overwhelmed by negative samples attained from the PA stage in the initial phase of training.
2000 RGB+D frames are extracted from our AID dataset and annotated with upper-body bounding boxes for training the PA stage.
Joints of interactors are also annotated in these 2000 frames.
Along with 3000 images selected from the MS-COCO dataset~\cite{coco}, a total of 5000 RGB images are used for training the Pose stage.

During the training of our AGANet, the model is optimized using Adam~\cite{adam} with the default parameter settings.
We train for 60 epochs with a batch size of 256.
For every epoch, subsequences with length T = 100 are sampled with a  stride of 5 frames from each complete sequence. 
A whole training process costs only $7 \sim 8$ minutes for AGANet with \textbf{111K} parameters.
During prediction, the temporal window with length T = 100 is slided with a stride of 20 frames.
All the following experiments conform to setups above.


\subsection{Efficiency of PAPNet}

We compare our PAPNet with CPN~\cite{topdown2} and OpenPose~\cite{paf}. 
These two methods stand for the latest and most effective methods for multi-person pose estimation, including top-down (CPN) and bottom-up (OpenPose) ones.
Some adjustments are made based on their original network structures:
For $v1$, we properly compress the network size, considering that only the joints of the upper body need to be estimated.
After pre-training on the MS-COCO dataset~\cite{coco}, we finetune them on our AID dataset.
For $v2$, we make further compression to focus on pose estimation in interaction scenarios while losing some generality to other scenes, and apply the same training data as our PAPNet.

As shown in Table I, our PAPNet achieves the best efficiency far ahead (8.3 $\times$ smaller and 2.4 $\times$ faster than the 2nd place) and competitive accuracy on a subset of 226 test images, with also outputs at higher resolution from the Pose stage.
Fig. 5 further shows the effects of Pre-Attention:
The model manages to adapt attention regions to human poses in diverse scenes, with robustness to position and scale changes of interactors caused by camera viewpoint movements (especially in Fig. 5(a),(b)), and eliminate interference from irrelevant people (in Fig. 5(c),(d)).

The following experiments for the action recognition level are all based on the skeleton data extracted by our PAPNet, except in Sec. III.F.

\begin{table}[h]
    \caption{
    Efficiency and accuracy of different methods for the interactor's pose estimation on the AID dataset.
    The fps are tested on NVIDIA Jetson AGX Xavier.
    }
    \begin{center}
        \begin{tabular}{c | c c c}
             \hline
             \hline
             \makebox[10mm]{model} &
             \makebox[15mm]{parameters} &
             \makebox[10mm]{fps} &
             \makebox[20mm]{PCK@0.15} \\
             \hline
             CPN v1~\cite{topdown2} & 46.0M & 33 & \textbf{97.31} \\
             CPN v2~\cite{topdown2} & 27.0M & 47 & 96.56 \\
             OpenPose v1~\cite{paf} & 42.0M & 15 & 96.70 \\
             OpenPose v2~\cite{paf} & 11.6M & 43 & 95.73\\
             \textbf{PAPNet} & \textbf{1.4M} & \textbf{112} & 96.00 \\
             \hline
             \hline
        \end{tabular}
    \end{center}
\end{table}

\begin{figure}[t]
    \begin{center}
        \includegraphics[width=1.0\linewidth]{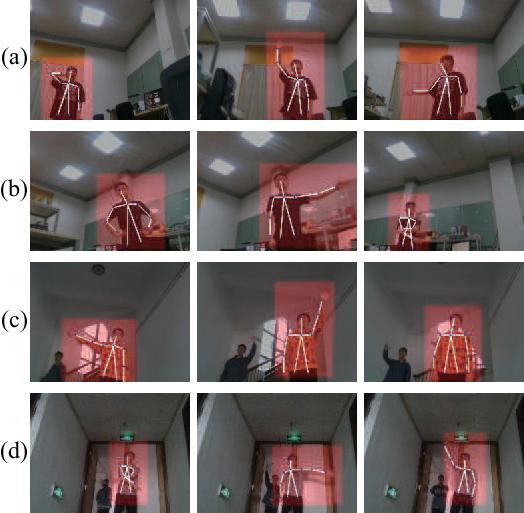}
    \end{center}
    \caption{
    \textbf{Examples of Pre-Attention and pose estimation results from PAPNet.}
    Local regions for Pre-Attention are masked out from the background.
    }
\end{figure}

\subsection{Effectiveness of base-AGANet}

We select several methods to compare with base-AGANet, the basic architecture of AGANet without LSTA or GSA modules.
These methods cover most of the popular network designs for encoding spatial-temporal patterns in skeleton-based action recognition:
(a) multi-layer CNN, a VGG-like deep CNN proposed in~\cite{sktcnn2}, 
(b \& c) multi-layer LSTM/biLSTM proposed in~\cite{sktrnn3, beyondjoints}, and
(d) multi-layer ST-GCN~\cite{stgcn}.
Neither do these methods import any attention-like mechanism.
Their detailed parameter configurations are adjusted to the size of AID dataset, and sizes of adjusted models are comparable to base-AGANet.
Prediction heads of them are also adjusted for dense frame-wise estimation.
Input sequences are arranged as skeleton images mentioned in Sec II. B, without any hand-crafted geometric features for a fair comparison.
These networks are all trained from scratch on AID dataset.

Table II shows the evaluation results. 
The 2nd lowest accuracy in the competition proves the multi-layer CNN to be unsuitable for our task.
As discussed in Sec. II. B, resizing a skeleton image with an unbalanced aspect ratio to typical image size results in deformations, and make some of the actions unrecognizable.
The 9.91 $cAP$ and 8.49 $AP_{trig}$ gap between biLSTM and LSTM shows the effects of backward information.
Besides, we find the two RNNs quickly falling into overfitting during training, possibly due to unnecessary encoding on over long time intervals in each layer.
The multi-layer ST-GCN achieves an effect close to base-AGANet (only 0.27 $cAP$ and 1.99 $AP_{trig}$ left behind).
However, its structure is composed of uniform blocks, which ignores distinction among information of different granularities.
As a comparison, different stages of base-AGANet focuses on patterns of different levels, thus better exploits the representation power of the network.
Furthermore, multi-level attention can be easily embedded in such a two-stage model.

\begin{table}[h]
    \caption{
    Recognition performance of different methods for skeleton-based action recognition on the AID dataset.
    }
    \begin{center}
        \begin{tabular}{c | c c c c}
             \hline
             \hline
             \makebox[15mm]{model} &
             \makebox[5mm]{$cAP$} &
             \makebox[8mm]{$AP_{trig}$} &
             \makebox[8mm]{$P_{trig}$} &
             \makebox[8mm]{$R_{trig}$} \\
             \hline
             multi-layer CNN~\cite{sktcnn2} & 75.30 & 78.41 & 78.46 & 79.39 \\
             multi-layer LSTM~\cite{sktrnn3} & 66.95 & 73.79 & 79.08 & 73.60 \\
             multi-layer biLSTM~\cite{beyondjoints} & 76.86 & 82.28 & 83.07 & 83.07 \\
             multi-layer ST-GCN~\cite{stgcn} & 81.63 & 87.56 & 87.72 & 87.41 \\
             \textbf{base-AGANet} & \textbf{81.90} & \textbf{89.55} & \textbf{90.61} & \textbf{88.74} \\
             \hline
             \hline
        \end{tabular}
    \end{center}
\end{table}

\subsection{Ablation study}

\textbf{Data augmentation strategies. }
As mentioned in Sec. III.C, we jointly implement three data augmentation strategies to enhance data diversity and avoid overfitting.
Various composites of the three proposed strategies are tested, as shown in Table III.
Each of these strategies benefits robustness and generalization performance of the model, while combined use of them achieves the best 8.44 and 7.38 increase in $cAP$ and $AP_{trig}$.

\begin{table}[h]
    \caption{
    Effects of data augmentation strategies in training.
    }
    \begin{center}
        \begin{tabular}{c c c | c c c c}
             \hline
             \hline
             \makebox[5mm]{rot} &
             \makebox[5mm]{dist} &
             \makebox[5mm]{gt} &
             \makebox[5mm]{$cAP$} &
             \makebox[8mm]{$AP_{trig}$} &
             \makebox[8mm]{$P_{trig}$} &
             \makebox[8mm]{$R_{trig}$} \\
             \hline
              &  &  & 79.06 & 88.62 & 89.00 & 86.11 \\
              &  & \checkmark & 81.84 & 91.38 & 91.43 & 89.11 \\
             \checkmark &  & \checkmark & 83.02 & 93.44 & 92.30 & 92.74 \\
              & \checkmark & \checkmark & 82.86 & 92.57 & 93.08 & 89.11 \\
             \textbf{\checkmark} & \textbf{\checkmark} & \textbf{\checkmark} & \textbf{87.50} & \textbf{96.00} & \textbf{95.08} & \textbf{95.71} \\ 
             \hline
             \hline
        \end{tabular}
    \end{center}
\end{table}

\begin{figure}[b]
    \begin{center}
        \includegraphics[width=0.85\linewidth]{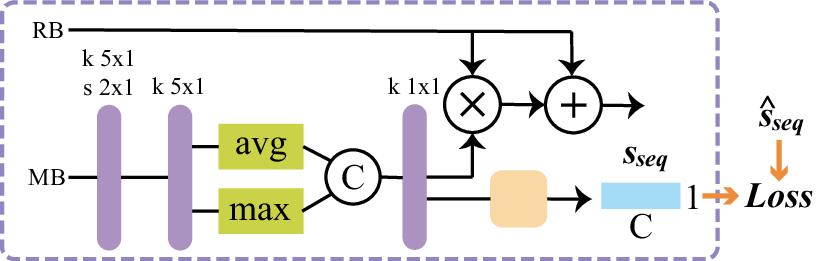}
    \end{center}
    \caption{
    \textbf{The soft mask branch of the GSA module with intermediate supervision appended at the end.}
    }
\end{figure}

\begin{figure*}[t]
    \begin{center}
        \includegraphics[width=0.9\linewidth]{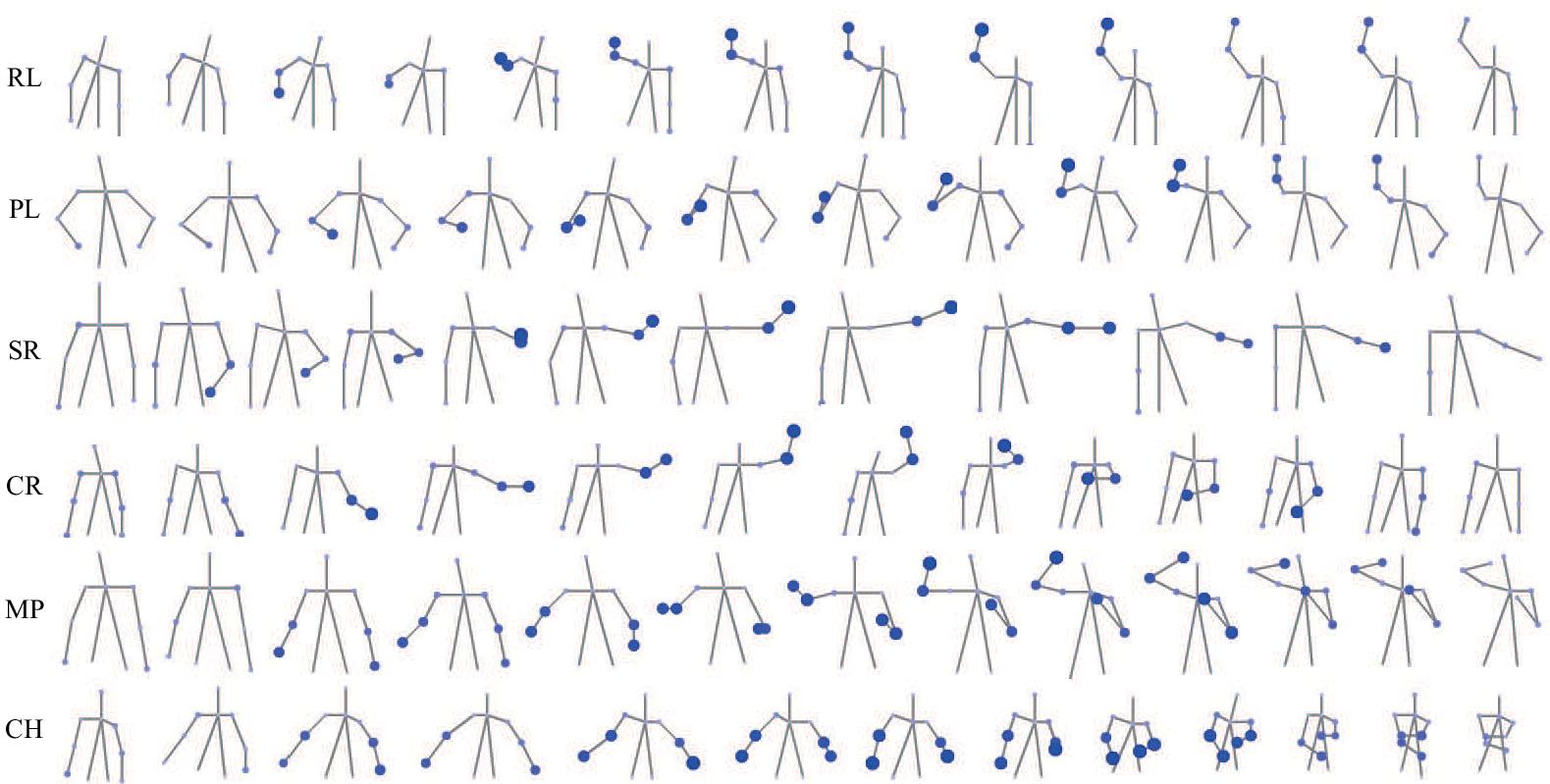}
    \end{center}
    \caption{
    \textbf{Examples of LSTA from AGANet on recognizing various actions.}
    Joints receiving more attention after LSTA are marked with larger and brighter circles. 
    }
\end{figure*}

\textbf{Attention modules in AGANet. }
Benefits from LSTA and GSA modules in our AGANet are evaluated.
From Table IV we can see that independently importing one of them improves the recognition results (3.71 $cAP$ and 4.23 $AP_{trig}$ increase by LSTA, 3.82 $cAP$ and 4.5 $AP_{trig}$ increase by GSA), and combined use of them also gives play to their respective advantages (totally 5.6 $cAP$ and 6.45 $AP_{trig}$ increase).
For further proof of GSA's effects in providing more comprehensive and precise semantic information, we append a regression layer at the end of the soft mask branch in the GSA module supervised by action categories $\widehat{s}_{seq}$ in the whole sequence, as shown in Fig. 6.
Such intermediate supervision (\textbf{imsp}) intends to explicitly guide the module to learn to capture global semantic information.
Additional \textbf{imsp} shows no advantages, which proves that our GSA module can capture global semantic information itself without explicit guidance.

\begin{table}[h]
    \caption{
    Effects of attention modules in AGANet.
    }
    \begin{center}
        \begin{tabular}{c c c | c c c c}
             \hline
             \hline
             \makebox[5mm]{LSTA} &
             \makebox[5mm]{GSA} &
             \makebox[5mm]{imsp} &
             \makebox[5mm]{$cAP$} &
             \makebox[8mm]{$AP_{trig}$} &
             \makebox[8mm]{$P_{trig}$} &
             \makebox[8mm]{$R_{trig}$} \\
             \hline
              &  &  & 81.90 & 89.55 & 90.61 & 88.74 \\
             \checkmark &  &  & 85.61 & 93.78 & 93.56 & 93.41 \\
              & \checkmark &  & 85.72 & 94.05 & 92.95 & 93.71 \\
             \checkmark & \checkmark &  & \textbf{87.50} & 96.00 & 95.08 & \textbf{95.71} \\
             \checkmark & \checkmark & \checkmark & 86.47 & \textbf{96.35} & \textbf{95.29} & 95.40 \\ 
             \hline
             \hline
        \end{tabular}
    \end{center}
\end{table}

We also visualize the attention distribution from the LSTA module while estimating certain sequences to give an intuitive impression of LSTA's effects.
As shown in Fig. 7, LSTA successfully conducts the model to focus on the main body parts involved in each action, e.g., left arm for RL/PL, right arm for SR/CR, and two arms for MP/CH.
Attention on these critical parts rises at the start of actions, maintains during the process and weakens at the end of actions.
Such a mechanism keeps in line with human intuition for perceiving others' actions in interaction.

\subsection{Sensitivity of AGANet to pose results}
We analyze the influence of pose estimation quality on AGANet's recognition performance.
Besides the PAPNet, we adopt two versions of CPN~\cite{topdown2} and OpenPose~\cite{paf} to provide skeleton data.
As Table V shows, there is no significant gap among recognition performance on different pose estimation results.
The analysis proves the robustness of AGANet to estimation errors from them.

\begin{table}[h]
    \caption{
    Recognition performance of AGANet based on different estimated pose results.
    }
    \begin{center}
        \begin{tabular}{c | c c c c}
            \hline
            \hline
            \makebox[25mm]{framework} &
            \makebox[5mm]{$cAP$} &
            \makebox[8mm]{$AP_{trig}$} &
            \makebox[8mm]{$P_{trig}$} &
            \makebox[8mm]{$R_{trig}$} \\
            \hline
            CPN v1 + AGANet & 88.47 & 96.56 & 96.02 & 94.39 \\
            CPN v2 + AGANet & 88.71 & 95.59 & 94.44 & 95.38 \\
            OpenPose v1 + AGANet & 86.13 & 95.81 & 95.93 & 93.40 \\
            OpenPose v2 + AGANet & 87.01 & 93.33 & 94.31 & 93.08 \\
            \textbf{PAPNet + AGANet} & 87.50 & 96.00 & 95.08 & 95.71 \\
            \hline
            \hline
        \end{tabular}
    \end{center}
\end{table}

\begin{figure}[t]
    \begin{center}
        \includegraphics[width=0.6\linewidth]{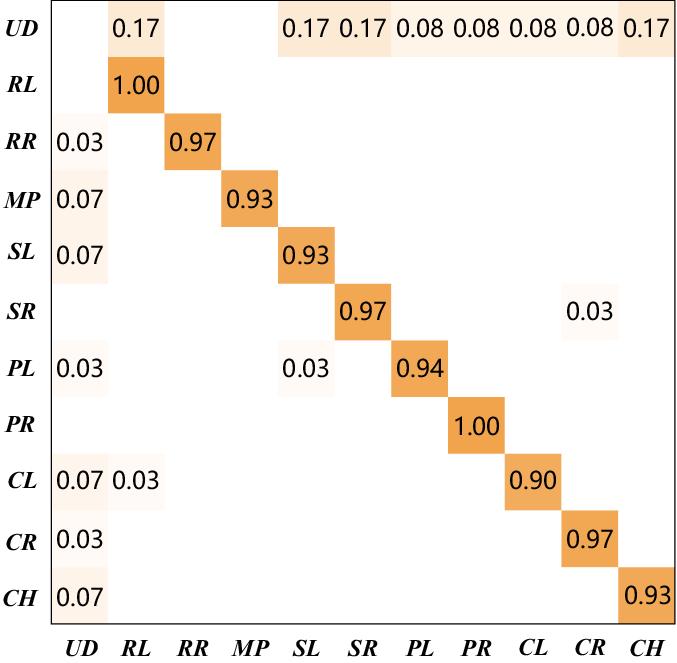}
    \end{center}
    \caption{
    \textbf{Confusion matrix of recognition with the whole framework on the AID dataset.}
    Vertical axis: groundtruth category.
    Horizontal axis: predicted category.
    "UD" denotes undefined actions.
    }
\end{figure}

\subsection{Error analysis}

As shown in Fig. 8, confusion mainly happens between defined actions and undefined actions.
We check the corresponding data and find that these false-positive instances have considerable similarities with defined actions, especially when observing the skeleton data.
A minimal amount of confusion between defined actions also originates from inter-category similarities, e.g., actions performed by the same limbs.
Overall, the proposed method has achieved satisfying results on the given task.
Extending action categories and tasks should provide better aids in HRI and we leave it for our future work.

\section{Conclusion}

In this work, we propose an attention-oriented multi-level network framework specifically for the action recognition task in HRI scenarios.
Compact architectures are designed at different levels for real-time interaction.
Furthermore, Pre-Attention employed in the pose estimation level manages to focus on the interactor and ensure the efficiency on mobile robot platforms.
LSTA and GSA modules incorporated in the action recognition level helps to capture important local structures and encode global semantic information.
Given promising performance on the newly constructed AID dataset, we believe that our approach can be extended to more complicated recognition tasks in HRI and facilitate further research in this field.

\bibliographystyle{IEEEtran}
\bibliography{my_paper}

\end{document}